\documentclass[runningheads]{llncs}
\usepackage[T1]{fontenc}
\usepackage{graphicx}
\usepackage{lipsum}
\usepackage{booktabs}
\usepackage{multirow}
\usepackage{subcaption}
\usepackage{xcolor}
\usepackage{hyperref}
\usepackage{enumitem}
\usepackage{amsmath}
\usepackage{cleveref}
\usepackage{amssymb}
\usepackage{xcolor}
\usepackage{comment}
\usepackage{booktabs}
\usepackage{siunitx}

\usepackage{orcidlink}

\begin{document}

\title{Anatomy-Aware 3D Mesh Refinement of Pericardium Segmentations on Computed Tomography}

% \title{Geometry-based refinement of a coarse automatic segmentation}

\titlerunning{Anatomy-Aware 3D Mesh Refinement}

\author{Andreas W. Aspe\inst{1}\orcidlink{0009-0003-2798-6586} \and Jonas Jalili Loft\inst{2}\orcidlink{0000-0002-9496-7655} \and Michael Huy Cuong Pham\inst{2}\orcidlink{0000-0002-6577-0079} \and Andreas Ohrt Johansen\inst{2}\orcidlink{0000-0002-7391-9945} \and Jørgen Tobias Kühl\inst{2}\orcidlink{0000-0002-7726-689X} \and Klaus Fuglsang Kofoed\inst{2, 3, 4}\orcidlink{0000-0001-9742-1554} \and Kristine Aavild Sørensen\inst{1,5}\orcidlink{0000-0003-1434-697X} \and Rasmus R. Paulsen\inst{1}\orcidlink{0000-0003-0647-3215} \and Josefine Vilsbøll Sundgaard\inst{1,5}\orcidlink{0000-0003-2872-4660}}

\authorrunning{A. W. Aspe et al.}

\renewcommand{\thefootnote}{\fnsymbol{footnote}}
\footnotetext[1]{~Code available at: \url{https://github.com/andreasaspe/3DMeshRefinement}}

\institute{DTU Compute, Technical University of Denmark, Kongens Lyngby, Denmark \email{awias@dtu.dk} \and Department of Cardiology, The Heart Center, Copenhagen University Hospital – Rigshospitalet, Copenhagen, Denmark \and Department of Radiology, The Diagnostic Center, Copenhagen University Hospital – Rigshospitalet, Copenhagen, Denmark \and Department of Clinical Medicine, Faculty of Health and Medical Sciences, University of Copenhagen, Copenhagen, Denmark \and Novo Nordisk A/S, Søborg, Denmark}

\maketitle

\begin{abstract}
Accurate delineation of the pericardium in a cardiac CT scan is essential for quantifying epicardial adipose tissue, yet it remains one of the most challenging structures to segment due to its poor contrast boundaries. Instead of solely relying on image gradients, our framework leverages the anatomical context of surrounding anatomical structures to guide the segmentation. This work introduces a novel 3D iterative mesh refinement framework that balances anatomical and geometric forces derived from inherent anatomical rules to refine an initial, possibly ambiguous, segmentation into a high-precision, anatomically plausible result. Designed as a model-agnostic post-processing step, our method uses a 3D vector field to iteratively push the vertices to the correct anatomical locations. Evaluating the refinement on both a high-resolution in-house dataset and a coarse, sparsely annotated open-source dataset, our method consistently improves all volumetric, surface, and anatomical metrics. The framework demonstrates greater improvement when applied to weaker initial segmentations, highlighting its potential for improving segmentations for out-of-domain models and in limited-training-data scenarios. The method is formulated as a gradient-based, GPU-accelerated framework that can be easily extended to other anatomical use cases.

%150-250 words

\end{abstract}

\section{Introduction}
Anatomical plausibility is often taken for granted in automatic medical image segmentation: state‑of‑the‑art architectures are expected to “implicitly learn” shape, topology, and organ relationships from the training data. In practice, this expectation is fragile. Thin, low‑contrast structures and inter‑subject variability mean networks can produce anatomically implausible contours, even when voxel‑wise metrics appear good.

The task of automated pericardium segmentation exemplifies these challenges. The pericardium is the outer sac of the heart, encapsulating all inner heart structures, and is routinely imaged using cardiac Computed Tomography (CT)~\cite{Wang2025_peri}. It appears on CT only as a thin, faint contour, and in some regions, it is barely visible. This makes the task remarkably difficult for both human annotators and automated methods, often risking overlap with surrounding anatomical structures due to poor contrast.

Accurate pericardial segmentation is critical for quantifying Epicardial Adipose Tissue (EAT), which is of high clinical interest due to its association with cardiometabolic diseases, including atrial fibrillation and heart failure~\cite{Pedersen2025,Iacobellis2022}. EAT lies inside the pericardial sac and can be isolated from the predicted pericardial mask on CT using a Hounsfield Unit (HU) threshold. Consequently, even small pericardial segmentation errors can remove significant EAT regions, yet remain minimally penalised by standard segmentation metrics. Existing automated methods are predominantly based on supervised deep learning paradigms, such as the U-Net architecture~\cite{Aspe2025}. While these models often achieve strong pixel-wise metric performance, they lack explicit anatomical constraints and may produce geometrically implausible segmentations.

To address this limitation, we propose an unsupervised, physics-inspired surface optimisation framework for pericardium refinement. Starting from a 3D mesh representation of an initial segmentation, we perform iterative deformation using a joint objective that balances anatomical plausibility with geometric surface regularisation. The anatomical forces are derived from a rich multi-organ anatomical context obtained from TotalSegmentator~\cite{Wasserthal2023}, which provides surrounding organs, tissues, and bones that collectively define a narrow spatial corridor in which the pericardium must reside. We exploit this anatomical confinement to guide the predicted surface toward a physically plausible configuration. Through gradient-based optimisation of the deformable mesh, implemented in a GPU-accelerated PyTorch3D framework~\cite{Ravi2020}, the vertices are iteratively adjusted to ensure that the final delineation is both accurate and anatomically consistent.

We evaluate our method on a local expert-annotated dataset and a public dataset using both quantitative and qualitative analysis. First, we demonstrate the effectiveness of the refinement using an open-source pericardium prediction model as a strong off-the-shelf initialisation. Second, we assess robustness across different initialisation models, including models trained with limited data and models applied in out-of-domain settings. Our contributions are:
\begin{enumerate}[label=(\alph*)]
    \item A novel, unsupervised 3D mesh refinement strategy that can be directly applied as a model-agnostic post-processing step for any initial pericardium segmentation.
    \item A tailored vector field and loss function that guide mesh vertex deformation using anatomical and geometric priors.
    \item A highly extensible, GPU-accelerated PyTorch3D implementation that is easily modifiable for other mesh refinement tasks, which can be guided by a vector field.
\end{enumerate}
\begin{figure}[t]
    \centering
    \includegraphics[width=\linewidth]{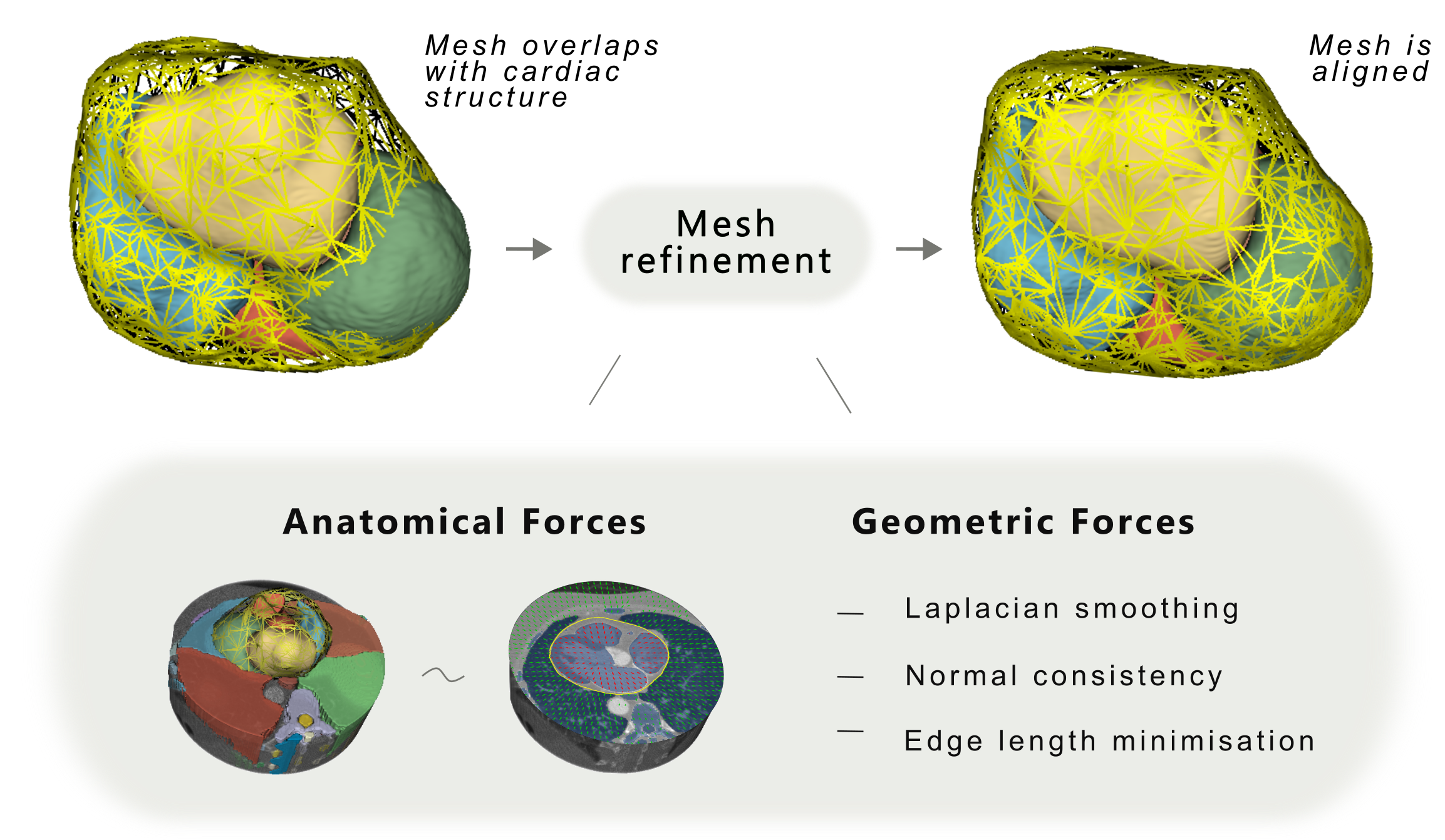}
    \caption{Overview of the iterative mesh‑refinement method. The initial pericardium mesh incorrectly intersects internal cardiac structures. By incorporating neighbouring anatomical masks as an anatomical force, and combining this with three geometric forces, the optimisation refines the initial mesh into an anatomically plausible mesh.}
    \label{fig:refinement}
\end{figure}
\subsection{Related Works}
In recent years, many research papers on automatic pericardium delineation have emerged. Early approaches often rely on traditional image analysis methods such as ellipse fitting~\cite{Rodrigues2017,Zlokolica2017} and atlas-based techniques~\cite{Norln2016}. Recently, this task has been predominantly addressed using U-Net architectures~\cite{He2020,West2023,Tang2024,Kuo2024}. TotalSegmentator~\cite{Wasserthal2023} recently included the pericardium in their open-source nnU-Net model~\cite{Isensee2021}, which has provided the community with an easily accessible baseline, enabling large-scale studies of EAT. While these U-Net-based models provide strong segmentation accuracy, they are optimised solely on a pixel-wise scale, inherently lacking global anatomical awareness.

To ensure anatomical plausibility, recent methods incorporate shape priors into deep learning. Oktay et al.~\cite{Oktay2018} introduced Anatomically Constrained Neural Networks (ACNNs) to penalise anatomically divergent predictions in the latent space. Alternatively, Painchaud et al.~\cite{Painchaud2020} developed a post-processing framework that uses a Constrained Variational Autoencoder (cVAE) to project invalid segmentations into a valid anatomical space.

To ensure geometric plausibility, classic methods have long relied on statistical shape modelling, balancing local image-driven forces with global shape constraints~\cite{Cootes1995,Zhang2013}. Li et al.~\cite{Li2021} refines an initial pericardium segmentation by balancing internal geometric smoothness with external image gradients. Recently, geometric priors have been integrated directly into deep learning frameworks with methods such as Deep Snake~\cite{Peng2020} and DeepSSM~\cite{Bhalodia2024}. Furthermore, approaches such as Voxel2Mesh~\cite{Wickramasinghe2020} demonstrate the efficacy of iteratively refining 3D meshes directly from volumetric data by minimising combined geometric loss functions.

While these existing methods successfully leverage either anatomical or geometrical priors, they typically address the target organ in isolation and rely heavily on specialised training data. We build upon this paradigm by proposing an unsupervised, 3D iterative mesh refinement method that explicitly combines both forces. In our framework, geometrical forces act as a shape prior for the balloon-like pericardium itself, while anatomical forces dynamically guide the refinement by enforcing strict boundary constraints based on all surrounding neighbouring organs.

\section{Methods}
\autoref{fig:refinement} illustrates the proposed mesh refinement framework. The figure demonstrates a common failure case in which the initial prediction of the mesh overlaps with internal cardiac structures, here the myocardium, resulting in an anatomically implausible segmentation. To correct the prediction, our refinement optimisation introduces an anatomical force that explicitly penalises these intersections. This force is derived from a vector field constructed from masks of neighbouring anatomical structures obtained using TotalSegmentator~\cite{Wasserthal2023}, an open-source whole-body AI segmentation tool. The resulting vector field guides the mesh toward anatomically valid boundaries defined by prior structural knowledge. At the same time, this guidance is balanced with three geometric forces - Laplacian smoothing, normal consistency, and edge length minimisation - to ensure a smooth and geometrically correct surface. All forces are balanced in a joint loss function and optimised iteratively using PyTorch3D~\cite{Ravi2020}, yielding a refined mesh that enforces strict spatial constraints while maintaining structural coherence.

\subsection{Construction of Anatomical Vector Field}\label{sec:vectorfields}
\begin{figure}[t]
    \begin{subfigure}{0.45\linewidth}
        \centering
        \includegraphics[width=\linewidth, angle=180]{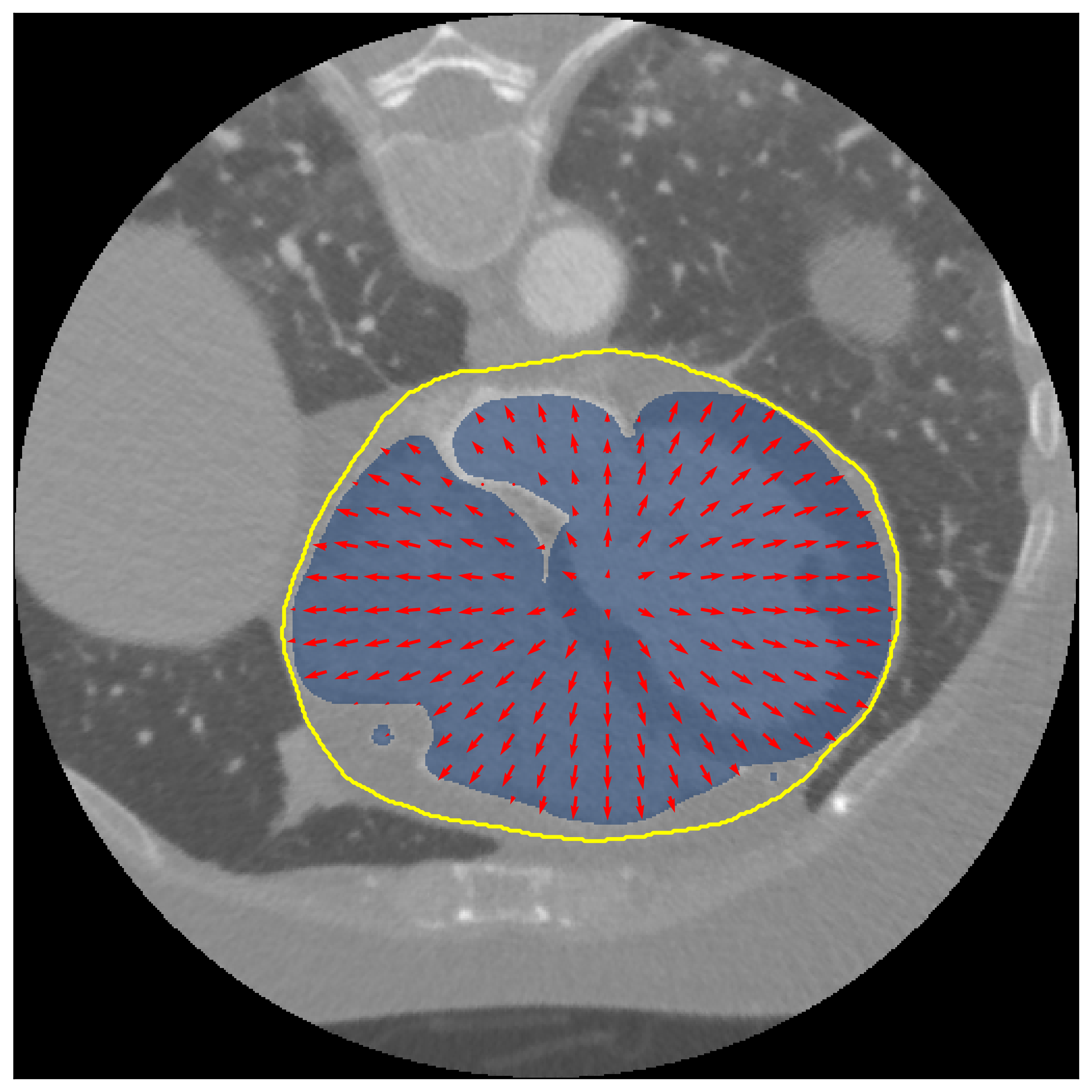}
        \caption{Internal vector field}
        \label{fig:vectorfield_inside}
    \end{subfigure}
    \hfill
    \begin{subfigure}{0.45\linewidth}
        \centering
        \includegraphics[width=\linewidth, angle=180]{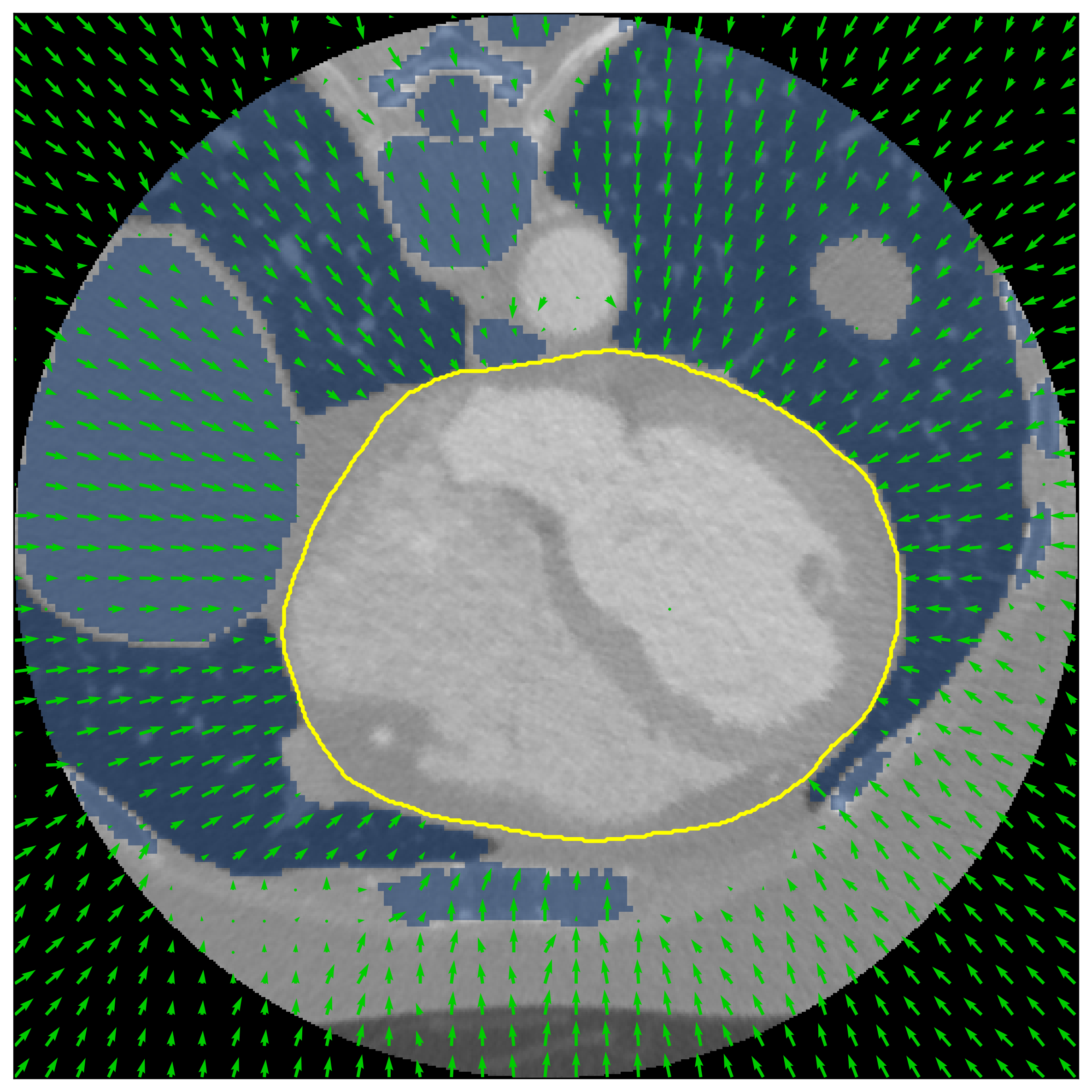}
        \caption{External vector field}
        \label{fig:vectorfield_outside}
    \end{subfigure}
    \caption{The internal and external vector fields derived from predicted anatomical masks of neighbouring organs. The yellow circle represents the outline of the ground truth pericardium.}
    \label{fig:vectorfield_figures}
\end{figure}

Based on established clinical knowledge~\cite{Arya2025}, we categorise surrounding anatomical structures as strictly 'internal' or 'external' relative to the pericardial sac. The heart chambers, myocardium, and coronary arteries are defined as internal, whereas the lungs, skeletal structures, and surrounding thoracic organs are external. These structures act as strict spatial boundaries that the mesh cannot intersect. To guide the mesh into the correct anatomical corridor, internal structures exert an outward repulsive force, while external structures exert an inward repulsive force. The resulting vector field is shown in~\autoref{fig:vectorfield_figures}, with the outward and inward force represented by red and green arrows, respectively. The vector field is carefully constructed to promote stable and accurate optimisation.

From anatomy, we know that the pericardium is a sac tightened around the internal heart structures. This physical topology dictates that the sac's surface normals align radially with respect to the centre of the heart. We use the Centre of Mass (COM), denoted as $\mathbf{c} \in \mathbb{R}^3$, of the heart chambers as our spatial reference.
For any spatial coordinate $\mathbf{x} \in \mathbb{R}^3$, we define a unit radial
reference direction $\mathbf{r}_{\text{com}}$: it points outwards for internal
structures ($\mathbf{r}_{\text{com}} = (\mathbf{x} - \mathbf{c}) /
\lVert \mathbf{x} - \mathbf{c} \rVert$) and inwards for external structures
($\mathbf{r}_{\text{com}} = (\mathbf{c} - \mathbf{x}) /
\lVert \mathbf{c} - \mathbf{x} \rVert$).
%For any spatial coordinate $\mathbf{x} \in \mathbb{R}^3$, we define a radial reference direction $\mathbf{r}_{\text{com}}$: it points outwards for internal structures ($\mathbf{r}_{\text{com}} = \mathbf{x} - \mathbf{c}$) and inwards for external structures ($\mathbf{r}_{\text{com}} = \mathbf{c} - \mathbf{x}$).

Let $\phi(\mathbf{x})$ denote the Signed Distance Field (SDF) of the anatomical mask, where $\phi(\mathbf{x}) \le 0$ inside the structure and $\phi(\mathbf{x}) > 0$ outside. To ensure that the mesh vertices are driven toward the correct anatomical corridor, we define the final force vector field $\mathbf{F}(\mathbf{x})$ as a single piece-wise function. 

Within the mask, the vectors are constrained to align strictly with the radial anatomical prior ($\mathbf{r}_{\text{com}}$) and are scaled by an exponential decay weight (with $\sigma = 2.0\,\si{\mm}$). This ensures that the applied forces gradually diminish as they approach the edge, preventing the vertices from overshooting. Outside the mask, the vectors follow the inverted normalised SDF gradient, $-\mathbf{g} = -\nabla\phi / \|\nabla\phi\|$, which inherently points toward the organ's surface from all surrounding directions. This creates a highly useful spatial dichotomy. On the side of an external structure furthest from the centre of the heart, these vectors naturally align with our radial prior ($\mathbf{r}_{\text{com}}$). Conversely, on the side nearest to the heart, they point in the exact opposite, undesired direction. To cleanly resolve this, we apply a soft stability gate that acts as a directional filter across both the internal and external vector fields. It preserves the vectors that align correctly with $\mathbf{r}_{\text{com}}$ and heavily dampens those pointing away. Consequently, if an initial mesh prediction extends to the far side of an organ, the vector field will still correctly guide it back toward the target corridor. The unified vector field is formulated as
\begin{equation}\label{eq:F(X)}
    \mathbf{F}(\mathbf{x}) = 
\begin{cases} 
\left( 1 - \exp\left(\frac{\phi(\mathbf{x})}{\sigma}\right) \right) \mathbf{r}_{\text{com}} & \text{if } \phi(\mathbf{x}) \le 0 \\ 
\max\left(0, -\mathbf{g} \cdot \mathbf{r}_{\text{com}}\right)^2 (-\mathbf{g}) & \text{if } \phi(\mathbf{x}) > 0 
\end{cases}
\end{equation}
\subsection{Loss Function}
\subsubsection{Anatomical Force}
The vector fields defined in~\autoref{eq:F(X)} drive the anatomical forces, pulling the mesh toward the target structural boundaries. To integrate these forces into a gradient-based optimisation framework, we must translate the desired spatial displacements into a differentiable loss function.

Let the mesh vertices be denoted by the set $V = \{\mathbf{v}_i\}_{i=1}^N$, where $N$ is the total number of vertices. Furthermore, let $\mathbf{F}: \mathbb{R}^3 \rightarrow \mathbb{R}^3$ represent the continuous vector field guiding the deformation.

At each iteration, the continuous vector field is sampled using trilinear interpolation to obtain a displacement vector $\mathbf{f}_i = \mathbf{F}(\mathbf{v}_i)$ for each vertex $\mathbf{v}_i$. The direction and magnitude of this sampled vector define the desired next location as $\hat{\mathbf{v}}_i = \mathbf{v}_i + \mathbf{f}_i$. We formulate the vector field loss as the Mean Squared Error (MSE) between the current vertex $v_i$ and the next optimal position $\hat{v}_i$
\begin{equation}
    \mathcal{L}_\text{vf} = \frac{1}{N} \sum_{i=1}^N \|\mathbf{v}_i - \hat{\mathbf{v}}_i\|_2^2
\end{equation}

By minimising the distance between the current vertex and target, the optimiser naturally pushes the vertices along the flow of the vector field, as the gradient simplifies to
\begin{equation}
    \nabla_{\mathbf{v}_i} \mathcal{L}_\text{vf}  \propto -\mathbf{F}(\mathbf{v}_i)
\end{equation}

Eventually, the vertices will reach the target boundary, where the vector field magnitude is zero. Consequently, the gradient of the vector field loss vanishes, and the vertices will stop moving unless pulled back into the anatomical mask by competing geometric forces, described in the next section, establishing a dynamic force balance. Importantly, $\hat{\mathbf{v}}_i$ is treated as a fixed spatial constant (stop-gradient), preventing gradients from flowing through $\hat{v}_i$ and the sampled vector field.
%A gradient descent-based optimiser takes a step in the negative direction of this gradient, leading to the vertices being systematically pushed along the vector field. 

\subsubsection{Geometric Force}
The anatomical force is counterbalanced by three geometric forces: Laplacian smoothing ($\mathcal{L}_\text{L}$), normal consistency ($\mathcal{L}_\text{N}$), and edge length regularisation ($\mathcal{L}_\text{E}$). These forces act as geometry priors and are the main driver in areas far from defined internal/external anatomies, where the vector field force is limited.

Laplacian smoothing minimises the distance between each vertex and the centroid of its immediate neighbours, inducing an elastic shrinkage. This effect causes the mesh to wrap tightly around the internal cardiac structures, accurately simulating the intricate, balloon-like geometry of the pericardium. Furthermore, the smoothing process improves overall mesh quality by mitigating vertex clustering and spiky artifacts.

Normal consistency enforces a globally smooth and anatomically plausible surface by penalising the cosine distance between the unit surface normals of adjacent triangular faces. This penalty ensures that the mesh does not exhibit abrupt angular variations or sharp creases during optimisation, which is consistent with the smooth, continuous nature of the pericardial surface.

Finally, we minimise the mean squared edge length with a target length of zero. This penalises excessively large triangles, helping to maintain a uniform mesh density and providing overall surface regularisation.

\subsubsection{Total Loss}
The overall objective function balances the anatomical with the geometric forces. To provide flexibility in weighting inward and outward deformations independently, the vector field forces are explicitly decoupled into internal ($\mathcal{L}_\text{vf-in}$) and external ($\mathcal{L}_\text{vf-ex}$) components. Therefore, the optimisation process minimises a weighted sum of five distinct components, yielding a total loss function formulated as
\begin{equation}\label{eq:loss}
    \mathcal{L}_\text{total} = \lambda_\text{in} \mathcal{L}_\text{vf-in} + \lambda_\text{ex} \mathcal{L}_\text{vf-ex} + \lambda_\text{E} \mathcal{L}_\text{E} + \lambda_\text{L} \mathcal{L}_\text{L} + \lambda_\text{N} \mathcal{L}_\text{N}
\end{equation}

The $\lambda$ hyperparameters control the relative influence of each force. By fine-tuning these weights, the optimisation process establishes a dynamic balance: the anatomical forces drive the mesh to conform precisely to the biological boundaries, while the geometric forces constrain the surface to remain smooth, cohesive, and physically plausible.

\section{Experiments}
\subsection{Datasets}
We evaluate our method using two CT datasets with ground truth pericardium annotations: a high-resolution, in-house annotated subset from the Copenhagen General Population Study (CGPS) and a low-resolution, open-source subset of the Sparsely Annotated Region and Organ Segmentation (SAROS) dataset. A comparison of the dataset characteristics is provided in Table \ref{tab:datasets}.
\subsubsection{CGPS}
The first dataset is a subset of CGPS~\cite{Fuchs2023}, a prospective cohort study of the general Danish population. The study includes randomly selected participants aged 40 to 100 years living in Copenhagen and the surrounding areas and comprises 13,183 CT scans as of March 2020. We utilise a random subset of 152 contrast-enhanced cardiac CTs from this larger cohort. The scans feature high-resolution voxel spacings of $\num{0.43} \times \num{0.43} \times \num{0.25}$ \si{\mm}. The pericardium was annotated by three medical experts using Vitrea 6.9 (Vital Images Inc., MN, USA). Initial contours were drawn approximately every 10th slice, followed by automatic software interpolation and manual correction across all three anatomical planes to ensure a complete volumetric ground truth.
\subsubsection{SAROS}
The second dataset is derived from SAROS~\cite{Koitka2024}, an open-source heterogeneous collection from The Cancer Imaging Archive (TCIA) that includes both non-contrast and contrast-enhanced scans. The full dataset consists of 900 CT scans, of which we selected a subset of 431 by making sure the heart was fully contained within the field of view without being cropped. This dataset features significantly lower resolution, with voxel spacings of $\num{0.84} \times \num{0.84} \times \num{5}$ \si{\mm}. Unlike the CGPS data, the ground truth pericardium in SAROS is annotated every 5th slice without interpolation, yielding approximately 4–5 annotated slices per volume. As a result, the annotations provide only a sparse sampling of the structure rather than a volumetric ground truth, in contrast to the densely annotated CGPS dataset.
\begin{table}[t]
\centering
\setlength{\tabcolsep}{12pt}
\caption{Comparison of Dataset Characteristics}
\label{tab:datasets}
\begin{tabular}{@{}lll@{}}
\toprule
Feature & CGPS & SAROS \\ \midrule
Scan Count & 152 & 431 \\
Gender (Women / Men) & 32\% / 68\% & 54\% / 46\% \\
Slice thickness & \SI{0.25}{\mm} & \SI{5}{\mm} \\
Annotation Type & Full Volumetric & Sparse (Every 5th slice) \\
\bottomrule
\end{tabular}
\end{table}
%Avg. Annotated Slices & $\approx 434$ & $\approx 4$ \\ 

\subsection{Experimental Design}\label{sec:experiment_design}
The experimental design consists of two objectives. The first objective is a quantitative and qualitative \textbf{performance benchmark} of the proposed refinement method using a strong off-the-shelf initialisation. For this, we utilise the newly released \texttt{trunk\_cavities} task from TotalSegmentator to predict the pericardium. Performance is evaluated on both CGPS and SAROS by comparing results before and after refinement against the ground truth. For each dataset, 10 scans are reserved for hyperparameter tuning, and the remaining scans are used for evaluation, resulting in 142 and 421 evaluation scans for CGPS and SAROS, respectively.

The second objective is to investigate how refinement performance depends on the quality of the initial segmentation, with a particular focus on \textbf{limited training data and domain shift} scenarios. This study is conducted using the CGPS dataset only. We evaluate seven different initialisation models, all predicting pericardium segmentations that are subsequently refined and evaluated on a fixed held-out test set of 52 CGPS cardiac CT scans. To study low-data regimes, we train five nnU-Net models on CGPS using subset sizes of 5, 10, 20, 40, and 100 scans. For out-of-domain evaluation, we include two externally trained models: the previously mentioned pericardium model by TotalSegmentator (referred to as `TotalSegmentator') and the model by Li et al.~\cite{Li2021}, distributed as the TIMESlice software. Due to deployment restrictions preventing direct use of TIMESlice on CGPS data, we construct a proxy model. TIMESlice is used to generate pseudo-labels for 1,000 cardiac CT scans from the open-source dataset ImageCAS~\cite{Zeng2023}, and a local nnU-Net is trained on these labels. This model is then used to predict initialisations on CGPS and is referred to as `TIMESlice*'.

\subsection{Evaluation Metrics}~\label{sec:evaluation_metrics}
\begin{figure}[t]
    \centering
    \begin{subfigure}{0.32\linewidth}
        \centering
        \includegraphics[width=0.8\linewidth]{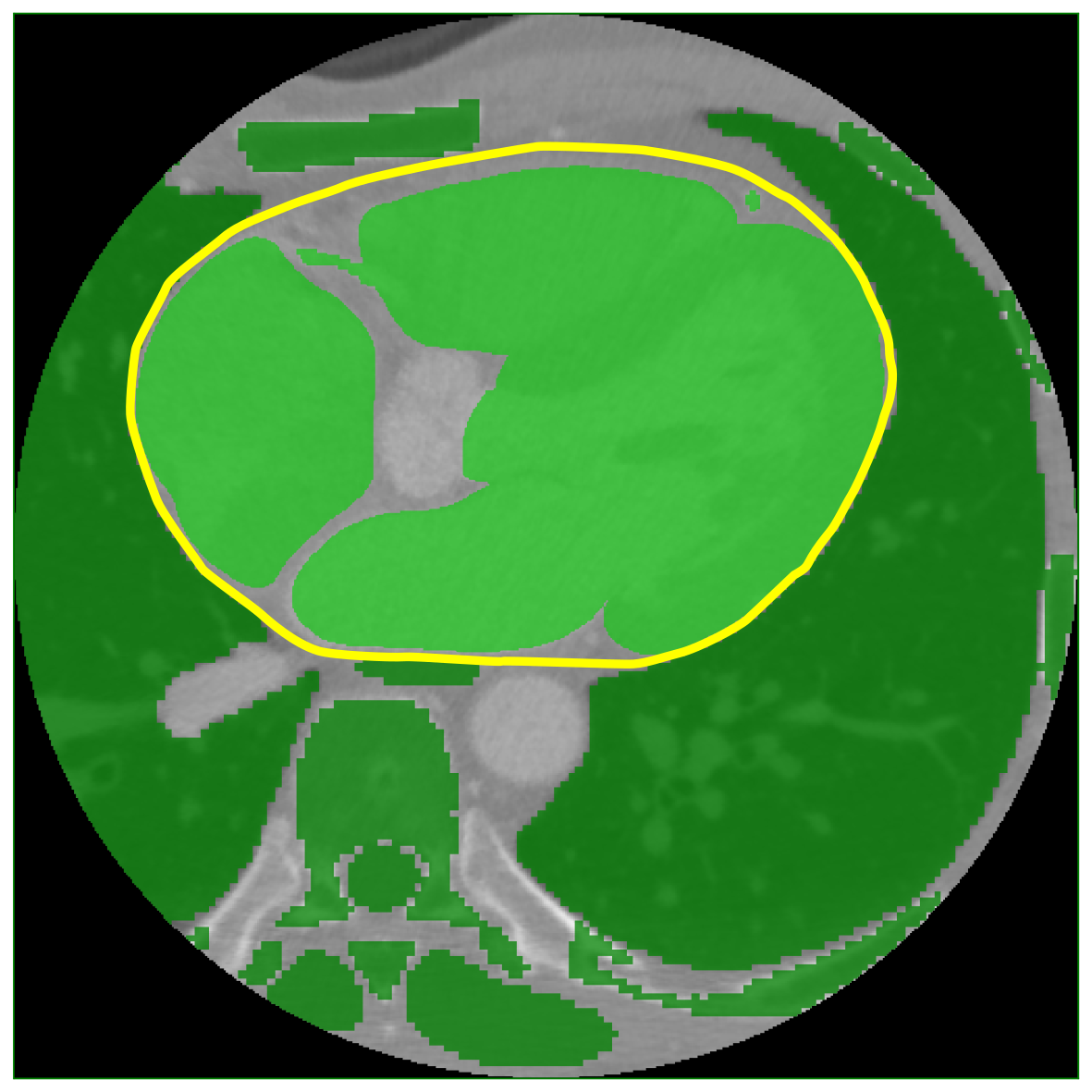}
        \caption{Ground truth}\label{fig:violation_outline}
    \end{subfigure}
    \hfill
    \begin{subfigure}{0.32\linewidth}
        \centering
        \includegraphics[width=0.8\linewidth]{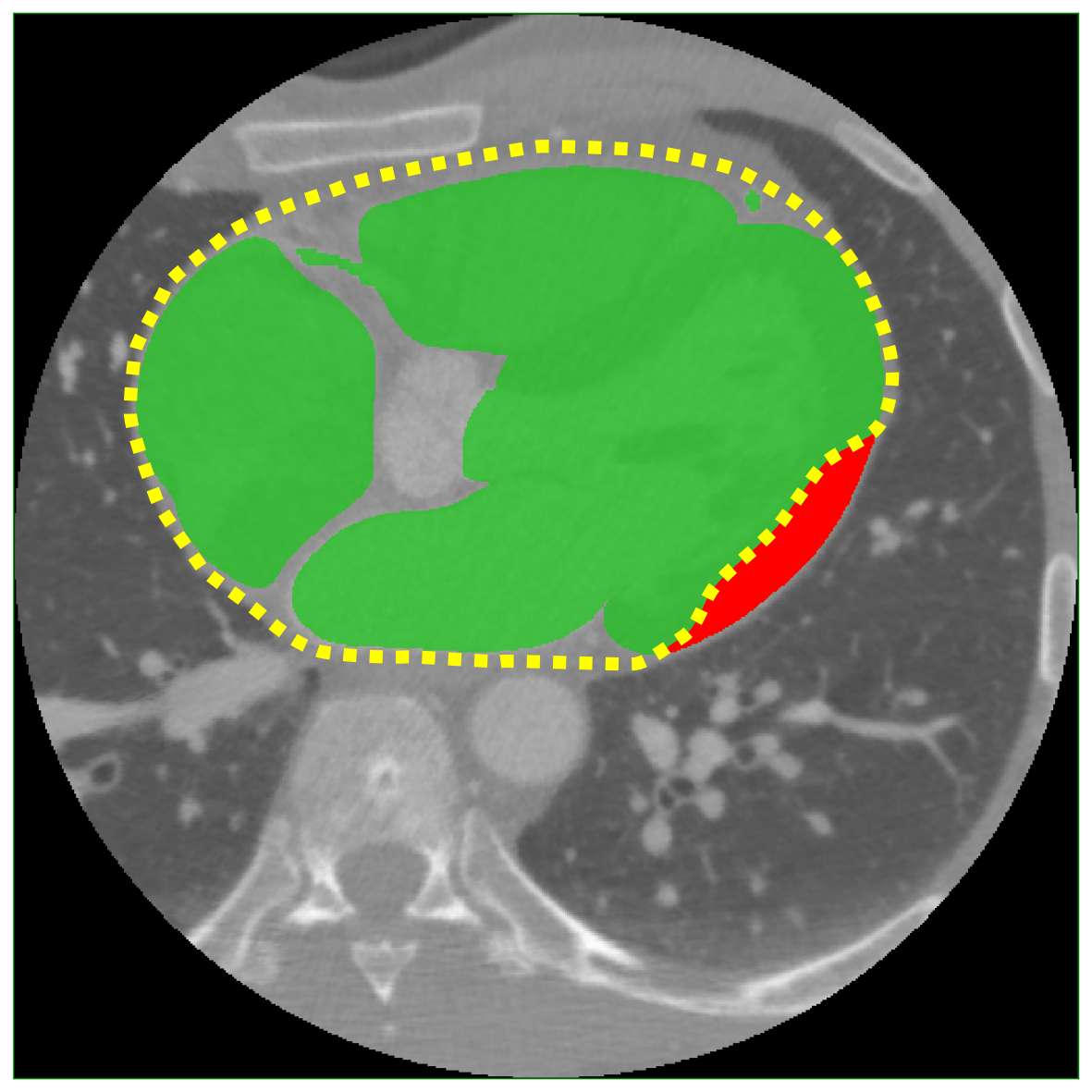}
        \caption{Internal violation}\label{fig:violation_internal}
    \end{subfigure}
    \hfill
    \begin{subfigure}{0.32\linewidth}
        \centering
        \includegraphics[width=0.8\linewidth]{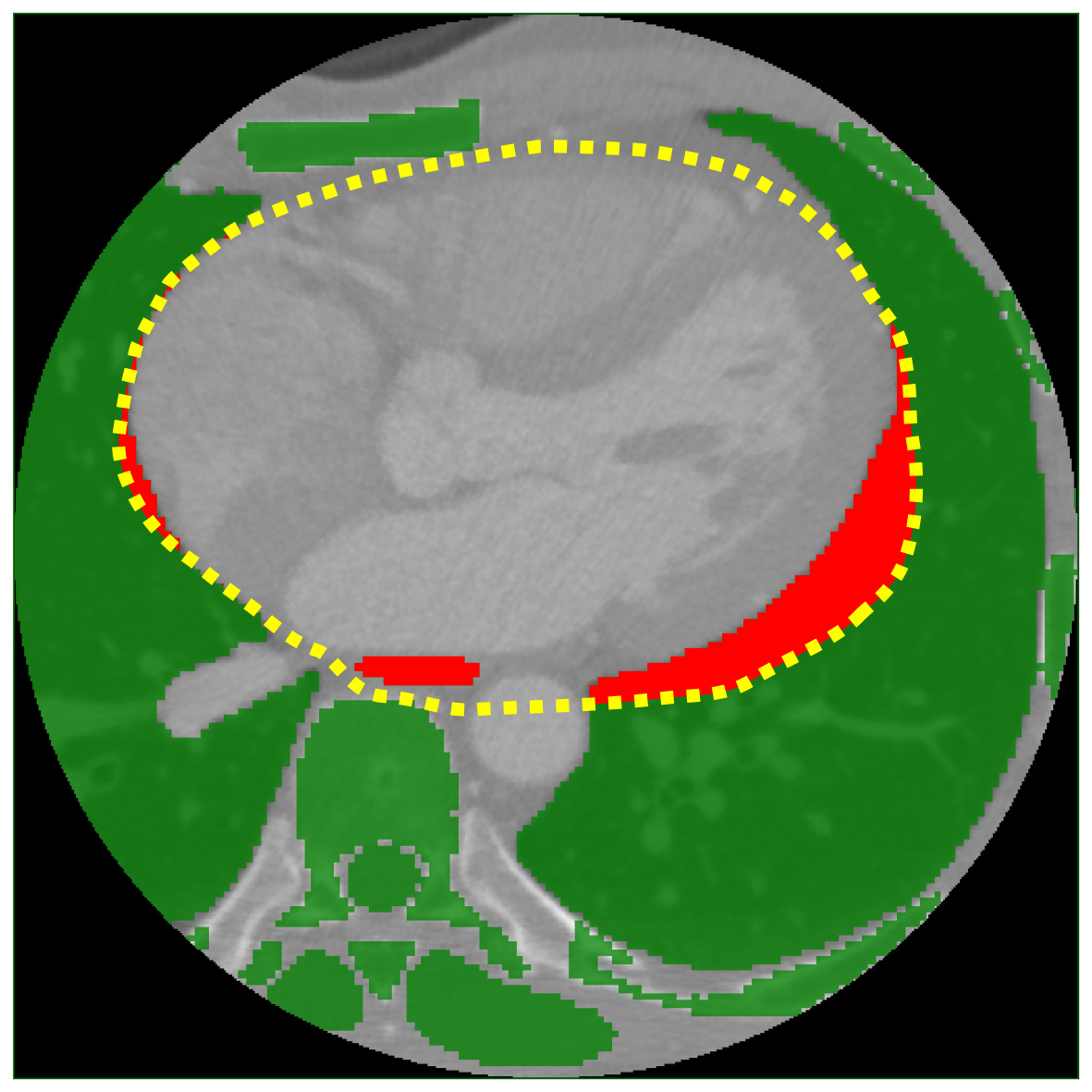}
        \caption{External violation}\label{fig:violation_external}
    \end{subfigure}
    \caption{Anatomically implausible segmentations. A predicted mesh (dotted yellow) may incorrectly exclude internal structures (internal violation) or encapsulate external structures (external violation). Violated tissues are highlighted in red relative to ground truth (solid yellow) and neighbouring anatomy (green).}
    \label{fig:anatomical_violations}
\end{figure}
To evaluate the refined pericardium segmentation, we compute the Dice Similarity Coefficient (DSC) for volumetric evaluation along with several boundary-based metrics: Average Symmetric Surface Distance (ASSD), 95th percentile Hausdorff Distance (HD95), and Normalised Surface Dice (NSD, $\tau = 1\ \si{mm}$). To assess the performance of the primary downstream task, we compute the DSC of EAT (denoted $\text{DSC}_{\text{EAT}}$), derived by thresholding the pericardial region (lower: -190 HU; upper: 0 HU)~\cite{Aspe2025,Xu2018}. We report all metrics as the mean across subjects along with the standard deviation.

Since there is no robust anatomical definition of the superior pericardial boundary, these metrics are computed only inferior to the top of the left atrium to ensure a fair comparison~\cite{Bartoli2022}. For the SAROS dataset, metrics are computed for each of the few annotated slices and averaged per subject; this slice-based evaluation may yield different metric characteristics compared to full 3D volumetric evaluation, as performed for CGPS.

Additionally, we quantify anatomical plausibility by measuring anatomical violations (\autoref{fig:anatomical_violations}). This metric calculates the volume of structures incorrectly excluded (internal violation) or falsely encapsulated (external violation) by the mesh relative to surrounding anatomical masks. These values are reported as median with interquartile range (IQR) as [Q1, Q3], as the distribution is non-normal due to high variability in violation magnitude across initial segmentations.

\subsection{Implementation Details}
The iterative mesh refinement is implemented as a GPU-accelerated optimisation in PyTorch3D~\cite{Ravi2020}. All code and hyperparameter settings are found on our GitHub: \url{https://github.com/andreasaspe/3DMeshRefinement}

Initial segmentations are obtained from voxel-wise segmentation networks and converted to a mesh using a marching cubes algorithm. The resulting surfaces are post-processed to obtain a compact and topologically consistent mesh suitable for optimisation, including connectivity filtering and smoothing operations.

Optimisation is performed in three stages by adjusting the relative weights
of the anatomical and geometric terms. The initial iterations prioritise anatomical
alignment through strong weighting of the vector field, while the second stage
gradually increases the emphasis on geometric regularisation to stabilise and
smooth the surface. In the third stage, the weights are held fixed at their final
values for the remaining iterations to allow for convergence, yielding a total of 2000 iterations. The optimisation is driven by the AdamW optimiser with a cosine annealing learning rate schedule. A light Taubin smoothing step is applied as final post-processing to reduce potential residual noise arising from unstable gradients.

\section{Results}
\subsection{Performance Benchmarking}
\label{sec:objective1}
\begin{table}[t]\centering
\caption{Segmentation performance before and after refinement. $\mathrm{\Delta}$ is the relative improvement. Improvements over the baseline are shown in bold.}
\setlength{\tabcolsep}{6pt}
\begin{tabular}{l lccc}
\toprule
Dataset & Metric & Initial mesh & Refined mesh & $\mathrm{\Delta}$ (\%) \\
\midrule
\multirow{7}{*}{CGPS}
& $\uparrow$ $\text{DSC}\, (\%)$ & $96.65 \pm 0.70$ & $\mathbf{96.81} \pm 0.65$ & $+0.16$\,\% \\
& $\uparrow$ $\text{DSC}_{\text{EAT}}\, (\%)$ & $85.09 \pm 3.41$ & $\mathbf{85.51} \pm 2.80$ & $+0.50$\,\% \\
& $\uparrow$ $\text{NSD}\, (\%)$ & $80.17 \pm 6.72$ & $\mathbf{81.74} \pm 6.11$ & $+1.96$\,\% \\
& $\downarrow$ $\text{HD95}\, (\text{mm})$ & $4.13 \pm 0.97$ & $\mathbf{3.91} \pm 0.96$ & $+5.20$\,\% \\
& $\downarrow$ $\text{ASSD}\, (\text{mm})$ & $1.19 \pm 0.26$ & $\mathbf{1.14} \pm 0.25$ & $+3.90$\,\% \\
& $\downarrow$ $\text{Internal Violation}\, (\text{cm}^3)$ & $1.05 \ [0.5, 1.8]$ & $\mathbf{0.00} \ [0.0, 0.0]$ & $+99.62$\,\% \\
& $\downarrow$ $\text{External Violation}\, (\text{cm}^3)$ & $0.89 \ [0.5, 1.9]$ & $\mathbf{0.26} \ [0.2, 0.4]$ & $+70.95$\,\% \\
\midrule
\multirow{7}{*}{SAROS}
& $\uparrow$ $\text{DSC}\, (\%)$ & $93.82 \pm 4.70$ & $\mathbf{94.29} \pm 4.34$ & $+0.49$\,\% \\
& $\uparrow$ $\text{DSC}_{\text{EAT}}\, (\%)$ & $84.21 \pm 5.99$ & $\mathbf{85.82} \pm 5.14$ & $+1.91$\,\% \\
& $\uparrow$ $\text{NSD}\, (\%)$ & $97.00 \pm 3.14$ & $\mathbf{97.22} \pm 2.88$ & $+0.23$\,\% \\
& $\downarrow$ $\text{HD95}\, (\text{mm})$ & $2.05 \pm 2.64$ & $\mathbf{1.88} \pm 2.60$ & $+8.36$\,\% \\
& $\downarrow$ $\text{ASSD}\, (\text{mm})$ & $0.26 \pm 0.48$ & $\mathbf{0.24} \pm 0.45$ & $+7.15$\,\% \\
& $\downarrow$ $\text{Internal Violation}\, (\text{cm}^3)$ & $1.22 \ [0.6, 2.8]$ & $\mathbf{0.02} \ [0.0, 0.1]$ & $+98.55$\,\% \\
& $\downarrow$ $\text{External Violation}\, (\text{cm}^3)$ & $0.55 \ [0.2, 1.3]$ & $\mathbf{0.43} \ [0.2, 0.9]$ & $+21.50$\,\% \\
\bottomrule
\end{tabular}
\label{tab:refinement_results}
\end{table}

\autoref{tab:refinement_results} summarises all metrics before and after refinement of the predicted pericardium segmentations from TotalSegmentator on the CGPS and SAROS datasets, as described in~\autoref{sec:experiment_design}. The refinement consistently improves all evaluated metrics. Improvements in pericardium DSC are modest, reflecting the already strong baseline performance, whereas the downstream EAT estimation shows larger relative gains, indicating that small geometric corrections of the pericardium have a stronger effect on EAT.

The most pronounced improvements are observed in boundary-based metrics (HD95 and ASSD), suggesting consistent fine-grained corrections of the pericardial surface. Standard deviations are generally lower for CGPS than for SAROS due to sparsity and variability in the latter dataset.

For anatomical plausibility metrics, refinement substantially reduces both internal and external violations. Internal violations are nearly eliminated, while a small amount of external overlap remains due to differences in the weighting of the internal and external vector fields. The smaller relative improvement in external violations for SAROS is consistent with its lower initial violation levels.

\begin{figure}[t]
    \centering
    \begin{subfigure}{0.47\linewidth}
        \centering
        \includegraphics[width=\linewidth]{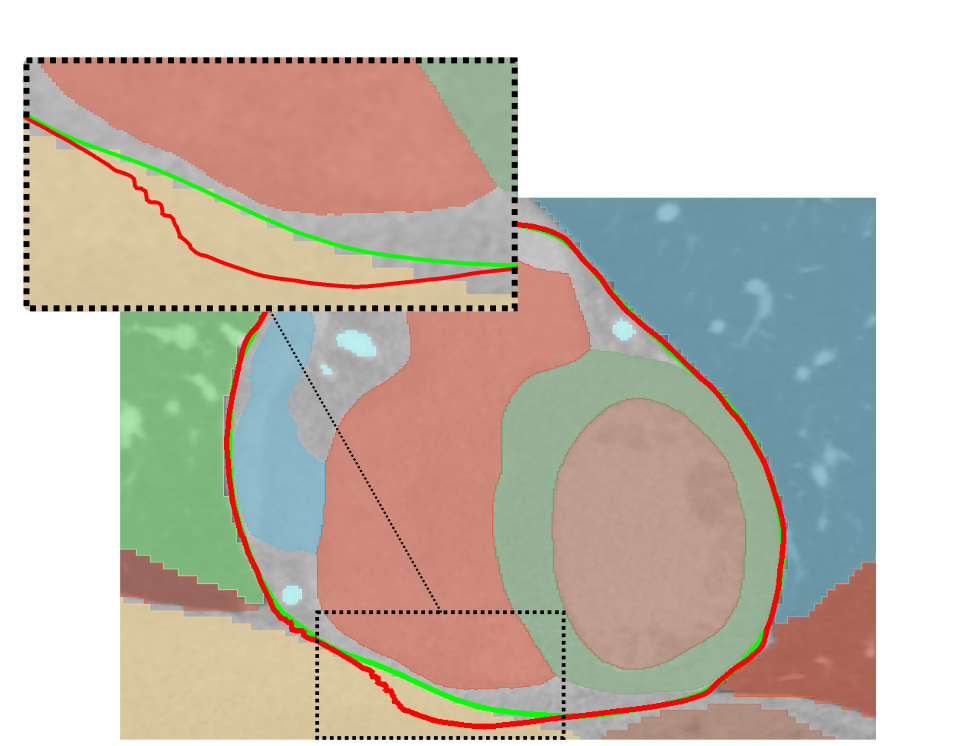}
        \caption{Overlap with liver}\label{fig:external_violation_corrected}
    \end{subfigure}
    % \hfill
    \begin{subfigure}{0.47\linewidth}
        \centering
        \includegraphics[width=\linewidth]{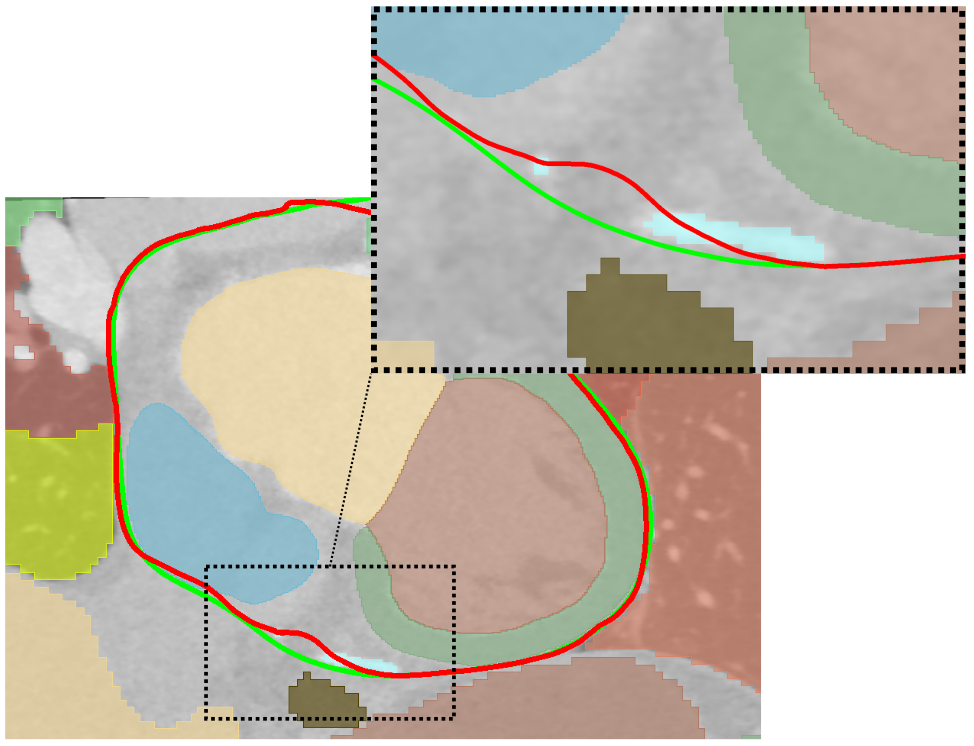}
        \caption{Overlap with coronary artery}\label{fig:coronary}
    \end{subfigure}
    \caption{Visual examples of anatomical corrections in the CGPS dataset. \textcolor{red}{Red} = initial mesh, \textcolor[rgb]{0.0,0.8,0.0}{Green} = refined mesh.}\label{fig:visual_corrections}
\end{figure}

The effect of the anatomical vector field is visually illustrated in~\autoref{fig:visual_corrections}. The figure shows two examples of coronal cross-sections from the CGPS dataset where refinement resolves external and internal violations, respectively. In~\autoref{fig:external_violation_corrected}, the red outline shows overlap with the liver, which is corrected by pushing the mesh inward to the green outline. Similarly, ~\autoref{fig:coronary} shows how the mesh is guided along the outer side of a coronary artery, correctly enclosing it beneath the pericardial surface while producing a smooth anatomically plausible surface.

Running 2000 iterations takes 9.8 s and 7.7 s for CGPS and SAROS, respectively, on an NVIDIA RTX PRO 6000 Blackwell Workstation Edition GPU.

%\subsubsection{Effect of refinement vs. initial model performance}
\subsection{Limited training data and domain shift analysis}\label{sec:objective2}
\begin{figure}[t]
    \centering
    \begin{subfigure}[t]{0.48\textwidth}
        \centering
        \includegraphics[width=\textwidth]{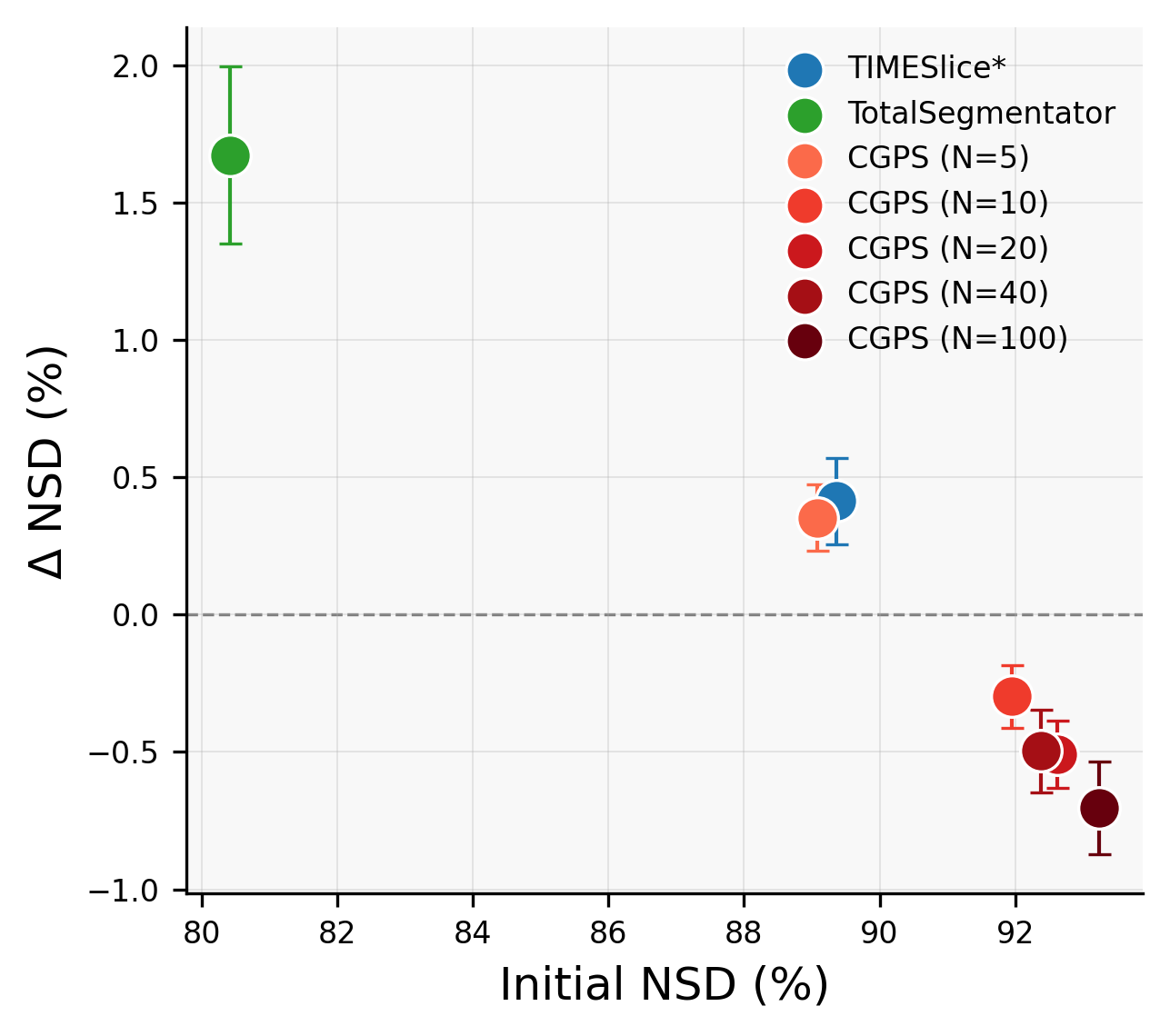}
        % \caption{DSC}
        \label{fig:DSC}
    \end{subfigure}
    \hfill
    \begin{subfigure}[t]{0.48\textwidth}
        \centering
        \includegraphics[width=\textwidth]{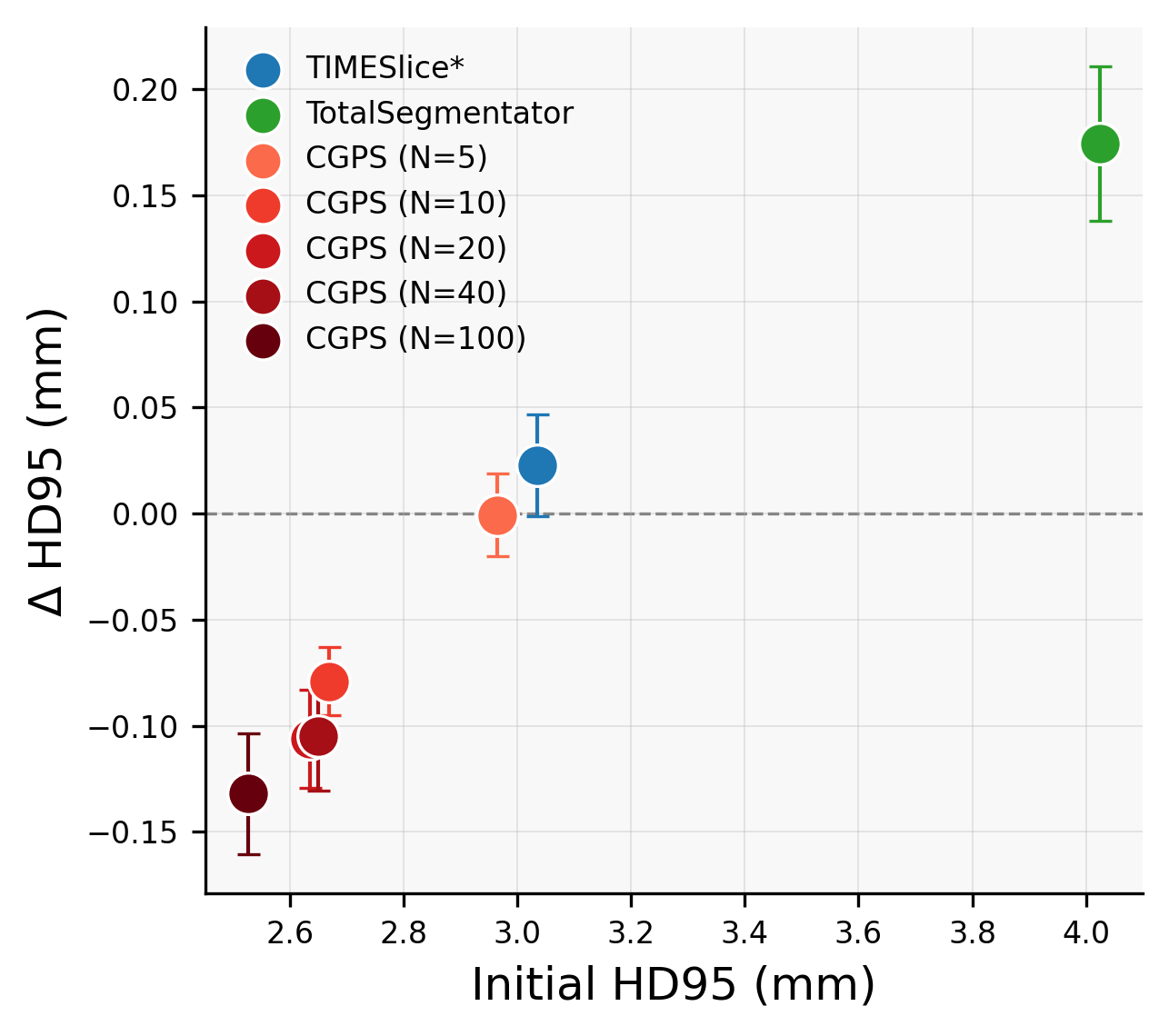}
        % \caption{NSD}\label{fig:NSD}
    \end{subfigure}
    \caption{Absolute improvement in NSD (a) and HD95 (b) as a function of initial performance across all baseline models. Results include CGPS models trained with $N \in \{5, 10, 20, 40, 100\}$ scans and two out-of-distribution models. Points above the dashed line ($y=0$) indicate a performance gain following refinement.}
    \label{fig:methods_comparison}
\end{figure}

\autoref{fig:methods_comparison} demonstrates a clear inverse relationship between initial segmentation quality and the absolute improvement after refinement for two surface metrics. In the extreme low-data regime ($N = 5$), refinement leads to an increase in NSD, while HD95 remains largely unchanged. For in-domain models trained with more data, a performance threshold emerges: when trained with $N \geq 10$ scans, refinement provides no consistent benefit and may instead lead to slight performance degradation.

Both out-of-domain models (TIMESlice* and TotalSegmentator) improve following refinement, with TotalSegmentator showing the largest gains due to its lower initial performance. Notably, TIMESlice*, which is trained on a substantially larger external dataset than the CGPS models, also benefits from refinement, indicating that the proposed method is particularly effective at mitigating domain shifts.

\section{Discussion}
We observe an inverse correlation between initial segmentation quality and refinement benefit: the lower the baseline performance, the greater the gain. This is explicitly evident in low-data scenarios and out-of-domain models. These findings highlight a critical practical guideline. If maximising accuracy on a specific target dataset is the priority, annotating a small number of scans (we found N > 10) to train a state-of-the-art architecture like nnU-Net remains the superior approach. If annotation resources are severely limited, training on just a few scans combined with our refinement approach provides a distinct advantage. Finally, if local annotation is entirely infeasible, applying our iterative mesh refinement to a model trained on publicly available data serves as a highly effective, unsupervised strategy for adapting segmentations to the target domain.

Our refinement improves the quality of pericardium segmentations by correcting local surface errors in pre-trained predictions using anatomical and geometric priors. The improvements are primarily reflected in surface-based metrics, suggesting that the method predominantly corrects local boundary irregularities rather than inducing substantial changes in global volume. In general, surface-based metrics are more appropriate for this task, as DSC is less sensitive to small boundary fluctuations that are particularly relevant for EAT estimation, which tends to accumulate near the pericardial boundary~\cite{Aspe2025}. This is further supported by the observation that the improvement in $\text{DSC}_{\text{EAT}}$ exceeds that of the pericardium itself.

The impact of the proposed refinement is particularly evident in the substantial reduction of both internal and external anatomical violations, highlighting its contribution to anatomical plausibility that is not captured by conventional metrics. As the anatomical metrics are directly influenced by the tuning of the vector field forces, a balance must be struck between anatomical constraints and geometric regularisation. The external vector field is generally coarser than the internal one, as it relies on anatomical context predicted by a lower-resolution model than that used for the internal structures; consequently, a higher degree of overlap is permitted to preserve a geometrically correct boundary. Importantly, correcting these violations improves the anatomical consistency of the segmentation and helps mitigate anatomically implausible inclusion of adipose tissue near the pericardial boundary, illustrating that even coarse AI‑generated segmentations of neighbouring organs can provide effective spatial guidance and enhance the robustness of the refinement framework.

The framework generalises to other anatomical refinement tasks where masks of neighbouring structures are available. Adapting it to new anatomies mainly requires defining an appropriate vector field within the current formulation, which is straightforward for large, smooth structures such as the liver or kidneys. Hyperparameters are largely transferable across different initialisation models and datasets with similar resolution, yielding consistent optimisation dynamics.

The method is computationally efficient, with mesh optimisation completing within a few seconds on a GPU. Because the computation relies on mesh coordinates rather than full 3D volumes, it maintains a minimal memory footprint, even for large numbers of vertices. Ultimately, this work serves as a plug-and-play correction module, which can be used as a post-processing step, rather than an add-on to already well-optimised in-domain segmentation models.

%Hyperparameters are expected to be largely transferable across datasets with similar spatial resolution, yielding consistent optimisation dynamics with minimal re-tuning.

\section{Conclusion}
This work introduces an unsupervised mesh refinement framework formulated as a gradient‑based, GPU‑accelerated optimisation that jointly balances anatomical and geometric forces. When applied to pre-trained pericardium predictions across two diverse CT datasets, the method consistently improves segmentation quality across all evaluated metrics. The refinement is most beneficial for weaker initial segmentations and offers diminishing or no returns for strong in-domain models, making it particularly useful in low-data settings or when relying on out-of-domain models. This framework ensures more anatomically plausible segmentations, increasing the accuracy of the downstream task of EAT estimation, without requiring additional annotations or retraining.

\begin{credits}
\subsubsection{\ackname} This study was supported by a research grant from Novo Nordisk A/S.

\subsubsection{\discintname}
The authors have no competing interests to declare that are relevant to the content of this article.
\end{credits}

\bibliographystyle{splncs04_6authors}
\bibliography{bibliography}

\end{document}